\documentclass{article}
\usepackage{arxiv}

\usepackage[utf8]{inputenc} 
\usepackage[T1]{fontenc}    
\usepackage{hyperref}       
\usepackage{url}            
\usepackage{booktabs}       
\usepackage{amsfonts}       
\usepackage{nicefrac}       
\usepackage{microtype}      
\usepackage{mathtools}
\usepackage{amsmath}
\usepackage{amssymb}
\usepackage{amsthm}
\usepackage{multirow}
\usepackage{graphicx}
\usepackage{textcomp}
\usepackage{color}
\DeclarePairedDelimiter\floor{\lfloor}{\rfloor}
\newtheorem{definition}{Definition}

\newcommand\blfootnote[1]{%
  \begingroup
  \renewcommand\thefootnote{}\footnote{#1}%
  \addtocounter{footnote}{-1}%
  \endgroup
}

\title{Stride and translation invariance in CNNs}

\author{
  Coenraad~Mouton, Johannes C.~Myburgh, Marelie H.~Davel \\
  Multilingual Speech Technologies, North-West University, South Africa; and CAIR, South Africa.\\
  \texttt{\{moutoncoenraad, christiaanmyburgh01\}@gmail.com} \\
}

\begin{document}
\maketitle
\blfootnote{This is a preprint - the final authenticated publication is available online at \url{https://doi.org/10.1007/978-3-030-66151-9_17}}
\begin{abstract}
Convolutional Neural Networks have become the standard for image classification tasks, however, these architectures are not invariant to translations of the input image.  This lack of invariance is attributed to the use of stride which ignores the sampling theorem, and fully connected layers which lack spatial reasoning.  We show that stride can greatly benefit translation invariance given that it is combined with sufficient similarity between neighbouring pixels, a characteristic which we refer to as \textit{local homogeneity}.  We also observe that this characteristic is dataset-specific and dictates the relationship between pooling kernel size and stride required for translation invariance.  Furthermore we find that a trade-off exists between generalization and translation invariance in the case of pooling kernel size, as larger kernel sizes lead to better invariance but poorer generalization.  Finally we explore the efficacy of other solutions proposed, namely global average pooling, anti-aliasing, and data augmentation, both empirically and through the lens of local homogeneity.
\end{abstract}


\section{Introduction}

\noindent Traditional computer vision tasks such as image classification and object detection have been revolutionized by the use of Convolutional Neural networks (CNNs) \cite{cnns}. CNNs are often assumed to be translation invariant, that is, classification ability is not influenced by shifts of the input image.
This is a desirable characteristic for image recognition, as a specific object or image must be correctly identified regardless of its location within the canvas area.  The assumption that CNNs exhibit translation invariance, however, has been shown to be erroneous by multiple authors \cite{subsampling_cnn,antialiasing_zhang,quantify_translation}, who all show that shifts of the input image can drastically alter network classification accuracy.  
Correcting for image translation within CNNs is an active area of study which has been partially addressed by methods such Spatial Transformer Networks \cite{spatial_transformer_networks}, Anti-Aliasing \cite{antialiasing_zhang}, and Global Average Pooling \cite{subsampling_cnn}.
 
In this work, we provide an overview of the properties that are required for translation invariance and show why they generally do {\it not} hold in the case of CNNs.  Furthermore we show that subsampling, commonly referred to as "stride" \cite{Goodfellow-et-al-2016}, can greatly benefit translation invariance.  However, we find that subsampling is only beneficial given that it is combined with sufficient similarity between neighbouring pixels, a characteristic that we refer to as \textit{local homogeneity}.     

We empirically explore the relation between local homogeneity and subsampling by measuring the effects of varying pooling kernel size and stride on both invariance and generalization, repeated across several different architectures and datasets.  We show that the inter-pixel variance of a given dataset dictates the necessary kernel size to ensure a degree of translation invariance.
In addition we examine the efficacy of solutions proposed by other authors, namely global average pooling, anti-aliasing, and data augmentation in terms of local homogeneity, translation invariance, and generalization.

Taken together, we clarify the various aspects that influence translation invariance in CNNs in order to better understand how certain architectural choices affect both invariance and generalization, and we further measure these effects to see if our hypotheses hold true.

\section{Background}

In order to understand translation invariance, we must first define the terms that play a role and their relation to one another.

\subsection{Translation Invariance and Equivariance}

\noindent Convolutional neural networks make use of convolution and pooling operators which are inherently translational, as filter kernels are shifted over an image to provide an output, commonly referred to as a feature map \cite{Goodfellow-et-al-2016}.  The translational nature of CNNs leads to the erroneous assumption that these systems are invariant to translations of the input image, or that the spatial location of internal features are irrelevant for classification.  However, Azulay and Weiss \cite{subsampling_cnn} show that even for state of the art CNN architectures (VGG16, ResNet50, InceptionResNetV2), a single pixel shift of the input image can cause a severe change in the prediction confidence of the network.  Surprisingly the shift of the input is imperceptible to the human eye, but drastically impedes the network's ability to classify.  

Translation invariance can be described as follows (as adapted from \cite{equivariance_equivalence}):
\begin{definition}
A function \textit{f} can be said to be invariant to a group of translations \textit{G} if for any \textit{g} element of G:
\begin{equation}
 f(g(I)) = f(I)
\end{equation}
\end{definition}
\noindent This implies that \textit{g} has no effect on the output of function \textit{f}, and the result remains equal.

A second misconception is that whilst modern CNNs are not translation \textit{invariant}, they are translation \textit{equivariant}.  Translation equivariance (also referred to as covariance by some authors) is the property by which internal feature maps are shifted in a one-to-one ratio along with shifts of the input. We define translation equivariance as follows (as adapted from \cite{equivariance_equivalence}): 

\begin{definition}
A function \textit{f} is equivariant to the translations of group \textit{G} if for any \textit{g} element of G:
\begin{equation}
    f(g(I)) = g(f(I))
\end{equation}
\end{definition}
This implies that the output is shifted in accordance with the shift of the input, or in other terms that the output of the function \textit{f} can be translated to produce the same result as translating the input \textit{I} before \textit{f} is applied would.

Intuitively it would be expected that translation equivariance holds for both convolution and pooling layers, and this intuition proves correct for dense convolution and pooling, if edge effects are ignored. To illustrate this property, an arbitrary one-dimensional filter is applied to a one-dimensional input, as well as to shifted values of this input.  
Consider an input signal \(I[n] = [0,0,0,0,1,2,0,0,0,0]\) and a kernel \(K[n] = [1,0,1]\), the result of the convolution \(I[n] \circledast K[n]\) is shown in the second column of Table \ref{tab:1d_equivariance_example}.

\begin{table}
\centering
\caption{Dense and strided convolution of a one-dimensional input and shifted variants}
\label{tab:1d_equivariance_example}
\begin{tabular}{c|c|c}
\textbf{Input}                                      & \textbf{Dense Convolution}                        & \textbf{Strided Convolution}             \\ 
\hline
\textcolor[rgb]{0.2,0.2,0.2}{[0,0,0,0,1,2,0,0,0,0]} & \textcolor[rgb]{0.2,0.2,0.2}{[0,0,1,2,1,2,0,0]}   & \textcolor[rgb]{0.2,0.2,0.2}{[0,1,1,0]}  \\
\textcolor[rgb]{0.2,0.2,0.2}{[0,0,0,0,0,1,2,0,0,0]} & {[}\textcolor[rgb]{0.2,0.2,0.2}{0,0,0,1,2,1,2,0]} & \textcolor[rgb]{0.2,0.2,0.2}{[0,0,2,2]}  \\
\textcolor[rgb]{0.2,0.2,0.2}{[0,0,0,0,0,0,1,2,0,0]} & \textcolor[rgb]{0.2,0.2,0.2}{[0,0,0,0,1,2,1,2]}   & \textcolor[rgb]{0.2,0.2,0.2}{[0,0,1,1]} 
\end{tabular}
\end{table}
As per this example, the equivariance property holds, as a shift of the input results in an equal shift of the output, meaning \(f(g(I)) = g(f(I))\).  However, this intuition fails when considering subsampling.

\subsection{Subsampling}

In CNNs subsampling occurs when a convolution or pooling layer is used with a kernel stride greater than one.  Using strided filters results in intermediary samples present in the input being skipped over and disregarded.  Strided filters are widely used in state-of-the-art architectures, which has both benefits and drawbacks.  

We first consider the disadvantages of subsampling. Azulay and Weiss~\cite{subsampling_cnn} correctly show that subsampling breaks translation equivariance, as the act of subsampling ignores the Nyquist-Shannon sampling theorem~\cite{nyquist}.  This theorem states that for a continuous signal to be fully reconstructed after sampling the sampling rate must be at least twice that of the highest frequency present in the input signal.  
In the case of CNNs, a digital input is sampled at a rate dictated by the kernel stride, implying that information is disregarded.  As information is lost, shifts of the input are not guaranteed to result in equivalent responses.  

This effect can be shown by using the previous example and a stride of 2 when calculating the convolution (thus a subsampling factor of 2) as the third column of Table \ref{tab:1d_equivariance_example} illustrates.

Two important characteristics of stride are illustrated by this example:
\begin{enumerate}
\item \textbf{Signal information is lost.} Compared to the output of dense convolution, it is clear that subsampling disregards intermediary values.  It is logical that this property can have an adverse effect on a CNN's ability to detect features in an image, as translating the input signal can cause features present in the input not to line up with the stride of the kernel.

\item \textbf{Translation Equivariance is lost.} The output of the convolution operation is no longer shifted in a one-to-one ratio with the input, meaning \(f(g(I)) \neq g(f(I))\), as can be seen by comparing the output of \(I[n] \circledast K[n]\)  to that of \(I[n-1] \circledast K[n]\).  This implies that shifts of the input signal result in outputs that are not equivalent.  
The result of the strided filtering of a translated input is not guaranteed to be sufficiently similar to that of its untranslated counterpart if equivariance does not hold; this partly explains why CNNs are so susceptible to small shifts.
\end{enumerate}

The main benefit of subsampling is that it can {\bf greatly reduce the training time} of CNNs.  As He and Sun \cite{time_complexity} show, the time complexity of convolution layers in a CNN is given by: 
\begin{equation}
O(\sum_{l=1}^d n_{l-1} \cdot s_l^2 \cdot n_l \cdot m_l^2)
\end{equation}
where $l$ is the index of the convolutional layer, $d$ is the number of convolutional layers, and for any layer $l$, $n_l$ is the number of output channels, $s_l$ the spatial size of the filter, and $m_l$ the size of the output feature map.
The subsequent output size of any given layer has a large effect, given that $m_l$ is squared.  It can also be shown that the output width of a layer $W_{l}$ given the previous layer width $W_{l-1}$ is specified by:
\begin{equation}
W_{l} = \frac{W_{l-1} - k_{w} + 2p}{s} + 1
\end{equation}
where \(k_{w}\) is the kernel width, \textit{p} is padding, and \textit{s} is stride (the same equation holds for height).  Given that stride acts as a divisor, we conclude that subsampling drastically reduces the size of the output, and in doing so substantially reduces the time complexity.  Furthermore, through spatial reduction, the size of the input layer to the fully connected layers is greatly reduced, implying there are less learn-able parameters which further reduces training time and memory consumption \cite{time_complexity}.
 
Whilst subsampling reduces spatial dimensions, it is worth noting that this can also be achieved by downsampling in CNNs.
We define downsampling as the reduction in spatial size caused by the size of a kernel during a convolution/pooling operation, where information along the edges of an input is disregarded (commonly referred to as ``edge effects").  Conversely, subsampling causes spatial reduction by explicitly disregarding intermediary samples.  Whilst downsampling does have an effect on translation invariance \cite{downsampling_spatial_location}, we attempt to mitigate this effect through the use of adequate padding.

\section{Signal Movement, Signal Similarity, and Local Homogeneity}

In this section we show how subsampling plays a role in terms of translation equivariance, and which characteristics must be accounted for when it is present.

\subsection{Shiftability}

Whilst subsampling breaks the equivariance property, we propose that it can greatly benefit translation invariance under certain circumstances due to a third property we define as shiftability.  Shiftability holds for systems that make use of subsampling and is defined as follows:

\begin{definition}
A function \textit{f} with subsampling factor \textit{s} is shiftable for a given translation if
\begin{equation}
f(g(I)) = g'(f(I))  
\end{equation}
where g is a translation function with translation vector \(\Vec{u}\) and input X
\begin{equation}
g(X) = t(\Vec{u}, X)  
\end{equation}
and
\begin{equation}
g'(X) = t(\frac{\Vec{u}}{s}, X)  
\end{equation}
\end{definition}
Put otherwise, shiftability holds for translations that are factors of the subsampling factor \textit{s} of \textit{f}.  When subsampling shifted inputs, equivalence will hold if a given translation is in accordance with the stride.  To illustrate this property, consider an arbitrary input signal that is subsampled by a factor of four and various shifts of the signal, as shown in Table \ref{tab:shiftability_example}.

\begin{table}
\centering
\caption{Subsampling factor of four for an arbitrary input}
\label{tab:shiftability_example}
\begin{tabular}{cc}
\multicolumn{1}{l}{Input}          & \multicolumn{1}{l}{0 0 0 0 0 0 0 0 0 3 2 5 2 4 1 6 3 4 6 5 5 0 0 0 0 0 0 0 0 0}  \\ 
\hline
\multicolumn{1}{l}{\textbf{Shift}} & \textbf{Subsampled Output}                                                       \\ 
\hline
0                                  & \textbf{0 0 0 2 3 5 0 0}                                                         \\
1                                  & 0 0 0 5 6 5 0 0                                                                  \\
2                                  & 0 0 0 2 1 6 0 0                                                                  \\
3                                  & 0 0 0 3 4 4 0 0                                                                  \\
4                                  & \textbf{0 0 0 0 2 3 5 0}                                                         \\
5                                  & 0 0 0 0 5 6 5 0                                                                  \\
6                                  & 0 0 0 0 2 1 6 0                                                                  \\
7                                  & 0 0 0 0 3 4 4 0                                                                  \\
8                                  & \textbf{0 0 0 0 0 2 3 5}                                                        
\end{tabular}
\end{table}

In this example, shiftability holds for translations that are factors of the subsampling factor (shifts of 4 and 8), and so a scaled form of equivariance is kept.  It is further evident that subsampling scales shifts of the input signal: in this example, a shift of four in the input results in only a shift of one in the output.

The subsampling factor dictates how many versions of the output signal can potentially exist after translation (again ignoring edge effects): In this example four discrete outputs are present, where all other outputs are merely shifted variants.  In the case of two-dimensional filtering, inputs are subsampled both vertically and horizontally, meaning \(s^2\) output signals can exist given a single input and a bounded translation.  This further implies that a given input will only be shiftable for \(\frac{1}{s^2}\) of possible translations \cite{subsampling_cnn}.

To explain how shiftability benefits translation invariance, we must first define two distinct characteristics that must be accounted for when comparing outputs of translated inputs to that of untranslated inputs.

\subsection{Signal Similarity and Signal Movement}

We propose two characteristics that influence translation invariance when subsampling is present:
\begin{itemize}
\item 
\textbf{Signal Similarity}: How much of the untranslated signal's output information is preserved after translation. 
\item
\textbf{Signal Movement}: How far the translated output has been moved from the original position of the untranslated output.
\end{itemize}

As an example, consider an output signal of [0,0,1,2,0,0] and another of [0,0,0,1,2,0].  These signals are exactly similar except that the second is shifted by one.  Conversely, an input of [0,0,1,2,0,0] and [0,0,2,3,0,0] are not exactly similar, but do not exhibit any shift (signal movement).

Due to the shiftability property it is evident that subsampling reduces signal movement, but leads to the loss of signal similarity, conversely using no subsampling leads to perfect signal similarity but allows the output to shift in a one-to-one ratio with the input.  For translation invariance in CNNs signal similarity must be kept but signal movement must also be reduced to the extent possible.
Whilst the reduction of signal movement during translation is fully dependent upon the subsampling factor, we propose that the degree of signal similarity that is preserved during subsampling is dependent upon \textit{local homogeneity}.

\subsection{Local Homogeneity}

As subsampling disregards intermediary pixels in a given feature map, translated versions will results in more equivalent outputs if neighbouring pixels are more similar to each other in a given region.  Put otherwise, preserving signal similarity requires that the variance between intermediary elements in a given window are sufficiently low enough, the size of which is dictated by the subsampling factor.  We call this property local homogeneity.  

To illustrate the effect of local homogeneity, consider an input that is fully locally homogeneous in accordance with the subsampling factor and the resulting output when shifting the signal.  This is shown in Table \ref{tab:signal_homogeneity} where a subsampling factor of 2 is used.
\begin{table}
\centering
\caption{Subsampling of a locally homogeneous signal}
\label{tab:signal_homogeneity}
\begin{tabular}{cc}
Input          & \multicolumn{1}{l}{0 0 0 0 0 0 2 2 3 3 1 1 2 2 0 0 0 0 0 0}  \\ 
\hline
\textbf{Shift} & \textbf{Subsampled Output}                                   \\ 
\hline
0              & 0 0 0 2 3 1 2 0 0                                            \\
1              & 0 0 0 2 3 1 2 0 0                                            \\
2              & 0 0 0 0 2 3 1 2 0                                            \\
3              & 0 0 0 0 2 3 1 2 0                                            \\
4              & 0 0 0 0 0 2 3 1 2                                           
\end{tabular}
\end{table}
Given that successive elements are similar in accordance with the subsampling factor, subsampling results in equivalent responses, yet the initial shift of the input is also scaled; implying that signal similarity is preserved and signal movement is negated.  Our hypothesis surrounding local homogeneity brings about two logical conclusions:
\begin{itemize}
\item 
\textbf{Pooling improves local homogeneity}: Dense pooling (max and average pooling alike) reduces the variance of any given input, therefore providing more similarity between neighbouring pixels.
\item
\textbf{Strided pooling can benefit translation invariance}: Combining a pooling window of sufficient size (and therefore sufficient local homogeneity) with subsampling leads to the reduction of signal movement.  This implies that translated samples will result in outputs that are more equivalent to untranslated samples.
\end{itemize}

In the following section we empirically explore our hypothesis surrounding local homogeneity by observing the effects of stride and pooling on translation invariance, as well as solutions proposed by other authors.

\section{Analysis}
In this section we empirically measure the translation invariance of different architectures and explore how these results relate to local homogeneity.
\subsection{Experimental Setup}
\subsubsection{Training} Two datasets are used for analysis, namely MNIST~\cite{MNIST} and CIFAR10~\cite{cifar}.
For prepossessing, MNIST networks are zero padded by 6 (meaning 6 rows of zeros are added to every edge of the image) and 10 for CIFAR10, resulting in 40x40 and 52x52 sized images respectively.  This allows space on the image canvas for translation.

We use cross-entropy loss and ReLU activation functions, along with the Adam optimizer \cite{adam}.  
Batch size is kept constant at 128 for MNIST networks, and a smaller batch size of 64 is used for CIFAR10 due to a lack of video memory to train exceptionally large networks.  Validation set size is constant at 5 000 for both MNIST and CIFAR10.

All MNIST networks are trained for a minimum of 100 epochs and 200 epochs for CIFAR10; furthermore, if a network has shown improvement in validation accuracy within the last 10 epochs an additional 15 epochs are added to training.  Step-wise learning rate decay is also used and the starting learning rate for a group of networks is chosen empirically, within the range of 0.001 to 0.0001.  It is confirmed that all networks are trained to 100\% train accuracy.

Post-training, the epoch exhibiting the highest validation accuracy is chosen to ensure the best generalization possible for the specific architecture. Apart from early stopping, we do not make use of use any explicit regularization methods such as dropout or batch-norm.

In terms of architectural choices, all convolution/pooling layers use padding to adjust for the downsampling caused by the size of the kernel, but not to mitigate the effects of subsampling, according to the following equation:
\begin{equation}
p = \floor{\frac{k_{w}-1}{2}}
\end{equation}

We use a standard three-layer architecture for both datasets, with 128 channels on the final layer and 3x3 dense convolution kernels, whilst pooling size and stride is varied for each experiment.
This architecture is useful as it performs relatively well on both datasets, given that no explicit regularization is used, but is still simple enough to interpret any subsequent results.

\subsubsection{Metrics} We measure translation invariance across different architectures by comparing the activation value vectors at the network output layer of an original and translated sample. Two measurements are used:
\begin{enumerate}
    \item 
Mean Cosine Similarity (MCS): As defined in Equation \ref{eq:mcs}, MCS uses the angle between any two vectors $a$ and $b$ as a similarity measure. Since only the angle is used, differences in vector magnitudes are ignored. 
\begin{equation}
\label{eq:mcs}
cos(\theta) = \frac{a \cdotp b}{\| a\| \| b\|}
\end{equation}
\item
Probability of top 1 change (PTop1): The probability of the top class prediction of a given network changing after an image is translated, as originally proposed by Azulay and Weiss~\cite{subsampling_cnn}.  This allows us to determine an exact probability of a sample being incorrectly classified given a range of translation.  This is useful as it purely measures a change in prediction accuracy and does not concern itself with other secondary effects.
\end{enumerate}

\subsubsection{Method of comparison} For each measurement, we report on the average value of the metric, across all test samples that are translated according to randomly sampling from a specified range, for both vertical and horizontal translation. This results in a single scalar value specifying the network's translation invariance for a given maximum range of translation.  Furthermore, when comparing several networks, only the samples that are correctly classified (before translation) by all networks present in the comparison are used.
Finally, all results are averaged over three training seeds. For results displayed with graphs, we include error bars indicating the standard error.

\subsection{Strided pooling and translation invariance}

\begin{figure}[h]
    \centering
    \includegraphics[scale=0.5]{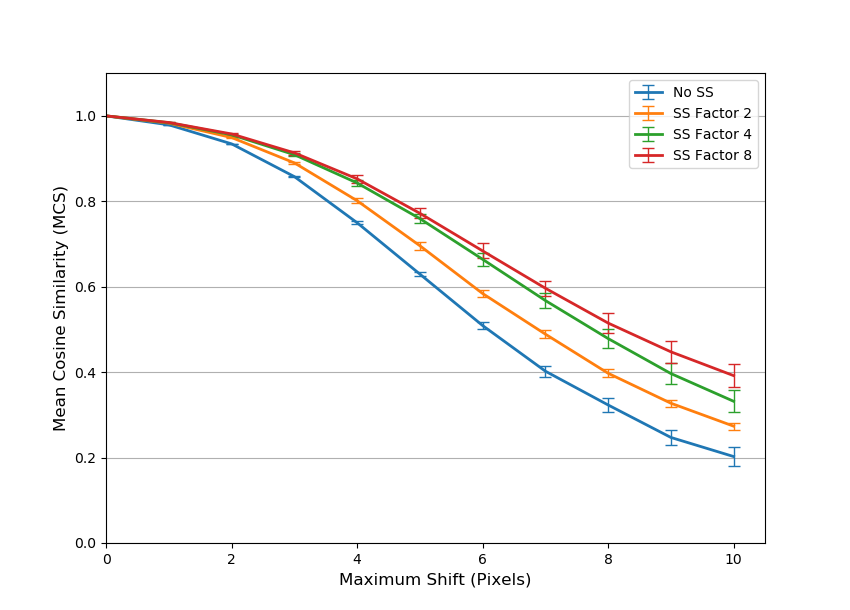}
    \caption{MCS comparison for MNIST architectures with varying subsampling}
    \label{fig:mnist_successive_ss}
\end{figure}

We explore the effects of strided pooling on translation invariance, using an architecture without subsampling as baseline.
Three other networks are trained with subsampling factors of 2, 4, and 8 respectively for the MNIST dataset.  The subsampling factor is varied by consecutively setting the stride of the 2x2 max pooling filters in the network to 2, starting at the first layer.  The comparative MCS is shown in Figure \ref{fig:mnist_successive_ss}.

\noindent This result shows that subsampling improves translation invariance for MNIST, where networks using subsampling show a substantially higher MCS.  This further implies that 2x2 max pooling is sufficient to provide local homogeneity given a stride of 2 for each layer, however samples in the MNIST dataset are generally inherently highly homogeneous. 

CIFAR10 is a more complex and detailed dataset, and is therefore generally less homogeneous and would require a larger degree of filtering to ensure translation invariance.  To ascertain the required max pooling size for this dataset, the previous experiment is repeated for kernel sizes ranging from 2x2 to 5x5 for each layer, as shown in Table \ref{tab:varysubsampling_varykernel_cifar}.
\begin{table}[h]
\centering
\caption{MCS for CIFAR10 networks with varying subsampling and max pooling kernel sizes (10 Pixel Range)}
\label{tab:varysubsampling_varykernel_cifar}
\begin{tabular}{c|l|l|l|l}
\multicolumn{1}{l}{~~\textbf{Subsampling Factor}} & \multicolumn{4}{c}{\textbf{Kernel Size} }                                                                 \\
\multicolumn{1}{l|}{}                             & \multicolumn{1}{c|}{2x2} & \multicolumn{1}{c|}{3x3} & \multicolumn{1}{c|}{4x4} & \multicolumn{1}{c}{5x5}  \\ 
\hline
1                                                 & 0.630                    & 0.598                    & 0.595                    & 0.618                    \\
2                                                 & 0.554                    & 0.635                    & 0.683                    & 0.731                    \\
4                                                 & 0.622                    & 0.674                    & 0.759                    & 0.789                    \\
8                                                 & 0.610                    & 0.660                    & 0.762                    & 0.791                   
\end{tabular}
\end{table}
We observe that in the case of CIFAR10 2x2 max pooling is not sufficient, and a substantial increase in translation invariance following subsampling is only observed at 3x3 pooling and larger.  

For networks that make use of subsampling, we observe that larger kernel sizes always result in greater invariance. Conversely, for networks that do not make use of subsampling (the first row of Table \ref{tab:varysubsampling_varykernel_cifar}) we observe a
a significant decrease in translation invariance as kernel-size is increased. Intuitively one would expect larger kernels to always provide greater translation invariance, but this intuition fails since these networks are fully translation equivariant.  
Finally, we observe that greater subsampling always results in greater invariance when adequately sized kernels are used (as in the case of 4x4 and 5x5 pooling) which are aligned with our findings on MNIST.

These results support our proposal that stride can significantly increase the translation invariance of a network, given that is combined with sufficient local homogeneity.  Furthermore we also find that the inherent homogeneity of a given dataset dictates the required filtering for subsampling to be effective.

\subsection{Anti-aliasing}
\label{sec:aa}
In terms of ensuring local homogeneity, average pooling outperforms that of max pooling, as an averaging kernel acts as an anti-aliasing (blurring) filter which greatly benefits translation invariance during subsampling.  However, Scherer et al.~\cite{Scherer2010EvaluationOP} show that max pooling results in better generalization than that of average pooling when used in CNNs.

Zhang~\cite{antialiasing_zhang} proposes a solution to this problem which allows the generalization benefits of max pooling without compromising translation invariance.  The author alters strided max pooling by separating it into two distinct layers:
(1) Dense Max Pooling, and 
(2) Strided Anti-Aliasing.
By applying an anti-aliasing filter, local homogeneity is ensured and the subsequent subsampling operation's affect on signal similarity is strongly mitigated, which results in a more translation invariant network.

The efficacy of this method is explored for both the MNIST and CIFAR10 datasets using the three layer 2x2 pooling networks from the previous section.  Each pooling layer present in the network is replaced with a dense max pooling layer and a bin-5 anti-aliasing filter.  These networks are then compared to their baseline counterparts that do not make use of anti-aliasing.  The comparative MCS for a maximum shift of 10 pixels is shown in Table \ref{tab:aa_cifar_mnist}.

\begin{table}[h]
\centering
\caption{Mean Cosine Similarity for MNIST and CIFAR10 networks with and without anti-aliasing for a maximum shift of 10 pixels}
\label{tab:aa_cifar_mnist}
\begin{tabular}{cccc}
\multicolumn{1}{l}{\textbf{Subsampling Factor} } & \multicolumn{1}{l}{\textbf{AA} } & \multicolumn{1}{l}{\textbf{MNIST} } & \multicolumn{1}{l}{\textbf{CIFAR} } \\ 
\hline
\multirow{2}{*}{1} & No & 0.248 & \textbf{0.630} \\
 & Yes & \textbf{0.329}  & 0.518 \\ 
\hline
\multirow{2}{*}{4} & No & 0.383 & 0.620 \\
 & Yes & \textbf{0.654}  & \textbf{0.710} \\ 
\hline
\multirow{2}{*}{8} & No & 0.447 & 0.611 \\
 & Yes & \textbf{0.638}  & \textbf{0.690}
\end{tabular}
\end{table}

For MNIST, anti-aliasing seems to always provide better translation invariance regardless of whether subsampling is used or not.  However, the greatest increase in translation invariance occurs when subsampling is applied, as it is not solely anti-aliasing that provides invariance, but its combination with stride.
For CIFAR10 the results are slightly different, anti-aliasing greatly reduces the invariance of the architecture without subsampling.  However, as is the case with MNIST, a large increase is evident when it is combined with subsampling.  

In conclusion, these results confirm that both signal similarity must be preserved and signal movement must be reduced to increase the network's invariance to translation, and that anti-aliasing is an effective solution for ensuring local homogeneity.

\subsection{No Subsampling and Global Average Pooling}

\noindent Whilst we have shown that the use of subsampling can greatly benefit translation invariance, Azulay and Weiss \cite{subsampling_cnn} propose a different approach that makes use of Global Average Pooling (GAP) and avoiding any subsampling throughout the network.  Global pooling is not influenced by signal movement, and with no subsampling, equivariance is kept, resulting in a perfectly translation invariant system.

We verify this by adding a final global average pooling layer to our baseline model without subsampling for CIFAR10, and we find that it has a 0\% Ptop1 change for shifts within the canvas area.  Put otherwise, the system is completely translation invariant.

Although this might seem to be a complete solution, GAP is not without its drawbacks.  Ignoring the benefits of subsampling, the GAP operation disregards a tremendous amount of information and could lower the classification ability of a given architecture, and is therefore not necessarily a suitable solution for every dataset.  However, Fully Convolutional Neural Networks (FCNNs) do make use of GAP, and withholding the use of subsampling in these architectures could be a suitable solution for ensuring translation invariance.

\subsection{Learned Invariance}

\begin{figure}[h]
    \centering
    \includegraphics[scale=0.7]{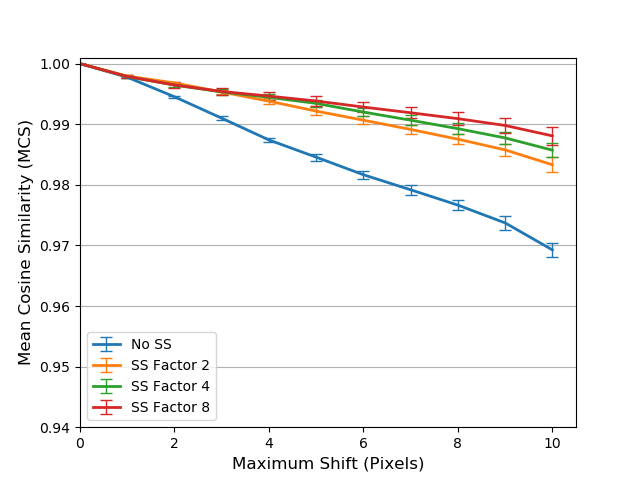}
    \caption{MCS comparison for MNIST architectures with data translation}
    \label{fig:mnist_successive_ss_translated}
\end{figure}

We explore the effects of learned invariance by training our previous MNIST architectures of Figure \ref{fig:mnist_successive_ss} on a translated data set.  We keep the size of the train set constant and randomly translate each sample up to a maximum of 8 pixels, furthermore we apply the same translation to the validation set (the test set remains untranslated).  In this way we explicitly optimize our models for translation invariance, and then examine whether our previous findings still hold true.  The comparative MCS of these networks is shown in Figure \ref{fig:mnist_successive_ss_translated}.

Observing this result, we find that the same pattern emerges as that of Figure \ref{fig:mnist_successive_ss}, where greater subsampling leads to greater translation invariance.  However, these networks are much more translation invariant than those not trained on translated data, with the lowest MCS at a staggering 0.97.

Whilst learned invariance is certainly a powerful tool, Azulay and Weiss point out that this method can potentially result in models that are overly biased to translations of the train set and it can not be expected to generalize well to translations of unseen data in all cases.  We also point out that MNIST is a particularly easy problem compared to more complex datasets such as CIFAR100 or ImageNet~\cite{ImageNet}, and usually data augmentation would be required for these networks to achieve good performance.  This implies that the training set must be explicitly augmented with translated data, which leads to a substantial increase in training time.

\subsection{Translation Invariance and Generalization}

\noindent We explore the relationship between translation invariance and generalization by comparing the test accuracy of our architectures to their MCS.  The test accuracy of our CIFAR10 networks of varying kernel size and subsampling from Table \ref{tab:varysubsampling_varykernel_cifar} is shown in Table \ref{tab:varysubsampling_varykernel_cifar_acc}.

\begin{table}[h]
\centering
\caption{Test accuracy for CIFAR10  networks  with  varying subsampling and kernel size}
\label{tab:varysubsampling_varykernel_cifar_acc}
\begin{tabular}{c|l|l|l|l}
\multicolumn{1}{l}{\textbf{Subsampling Factor}~ } & \multicolumn{4}{c}{\textbf{Kernel Size} } \\
\multicolumn{1}{l|}{} & \multicolumn{1}{c|}{2x2} & \multicolumn{1}{c|}{3x3} & \multicolumn{1}{c|}{4x4} & \multicolumn{1}{c}{5x5} \\ 
\hline
1 & 72.33 & 75.00 & 76.10 & 76.00 \\
2 & 74.43 & 77.00 & 77.57 & 76.69 \\
4 & 73.94 & 76.72 & 77.25 & 76.76 \\
8 & 72.53 & 75.31 & 76.69 & 75.95
\end{tabular}
\end{table}

\noindent We observe that larger kernel sizes generally generalize better, but also that kernels that are too large (such as 5x5 in this case) lead to a reduction in test set accuracy.  This is an expected result - larger kernels lower the variance of a given sample and result in more locally homogeneous regions, but also implies that more information is disregarded which negatively impacts the model's ability to generalize to samples not seen during training. Similarly, some subsampling seems to always provide better generalization regardless of kernel size, but too much subsampling leads to a reduction in model performance.  These differences suggest that there is a slight trade-off between a model's inherent invariance to translation and its generalization ability.

For the anti-aliased models of section 4.3 we observe a very small overall effect on generalization: Table \ref{tab:aa_test_accuracy} shows the test accuracy of these models with and without the use of anti-aliasing filters.

\begin{table}[h]
\centering
\caption{MNIST and CIFAR10 test accuracy with and without anti-aliasing}
\label{tab:aa_test_accuracy}
\begin{tabular}{clcc}
\multicolumn{1}{l}{\textbf{Subsampling Factor} } & \textbf{AA}  & \multicolumn{1}{l}{\textbf{MNIST} } & \multicolumn{1}{l}{\textbf{CIFAR10} } \\ 
\hline
\multirow{2}{*}{1} & No & \textbf{99.36}  & 72.33 \\
 & Yes & 99.18 & \textbf{73.62} \\ 
\hline
\multirow{2}{*}{4} & No & \textbf{99.38}  & 73.94 \\
 & Yes & 99.35 & \textbf{74.71} \\ 
\hline
\multirow{2}{*}{8} & No & \textbf{99.3}  & 72.53 \\
 & Yes & 99.21 & \textbf{73.40}
\end{tabular}
\end{table}

The combination of subsampling with anti-aliasing actually improves generalization for the CIFAR10 dataset, and only slightly hampers accuracy for that of MNIST.  These results are aligned with that of Zhang which show a slight improvement in generalization for state-of-the-art ImageNet networks using anti-aliasing.
These results, along with those of Section \ref{sec:aa}, show that for these data sets anti-aliasing is effective at improving translation invariance without reducing generalization ability.

\section{Conclusion}

We investigated the effect of stride and filtering on translation invariance and generalization in CNNs.  Our main findings are summarised below:

\begin{itemize}
\item 
Subsampling can greatly benefit translation invariance, given that it is combined with local homogeneity.  If sufficient filtering is used, greater subsampling leads to greater translation invariance.
\item
The amount of filtering required depends on the inherent homogeneity of a given data set.
\item
Too much filtering or subsampling negatively affects generalization, as such, a trade-off exists between translation invariance and generalization.
\item
Anti-Aliasing performs well in ensuring local homogeneity and therefore greatly increases an architecture's invariance to translation; it also performs better than large max pooling kernels in terms of generalization.
\item
We find that data translation, or learned invariance, is very effective, but can not be expected to perform well in all cases.
\end{itemize}

In terms of further research, we do not yet measure the effects of strided convolution, which could also be beneficial to translation invariance, nor do we quantify the effects of downsampling.  Along with this, several other methods that have been proposed, such as Spatial Transformer Networks~\cite{spatial_transformer_networks} and Capsule Networks~\cite{capsule} also require further exploration.

\bibliographystyle{unsrt}  
\bibliography{references} 

\end{document}